\pdfoutput=1
\documentclass{article}[draft]
\usepackage{subcaption}
\usepackage{titling}
\usepackage{graphicx}
\usepackage[table, xcdraw]{xcolor}
\usepackage{svg}
\usepackage{amsmath, amssymb}
\usepackage[utf8]{inputenc}
\usepackage{multicol}
\usepackage[top=25mm,bottom=30mm,inner=20mm,outer=35mm]{geometry}
\usepackage{mathptmx}
\usepackage[linecolor=red,backgroundcolor=red!25,bordercolor=red,textwidth=30mm]{todonotes}
\usepackage{blindtext}
\usepackage[font=small, labelfont=bf]{caption}
\usepackage{authblk}
\usepackage{xr}
\usepackage[section]{placeins}
\usepackage[colorlinks,linkcolor=red,anchorcolor=black,citecolor=green]{hyperref}
\usepackage{abstract}
\usepackage{cite}

\usepackage{soul}
\usepackage{boldline}
\definecolor{lightgray}{gray}{0.9}
\usepackage{algorithm}
\usepackage[noend]{algpseudocode}
\usepackage{hyperref}
\usepackage{cleveref}
\usepackage{comment}

\title{Advancing diagnostic performance and clinical usability of neural networks via adversarial training and dual batch normalization}

\author[1]{Tianyu Han}
\author[2]{Sven Nebelung}
\author[3]{Federico Pedersoli}
\author[3]{Markus Zimmermann}
\author[3]{Maximilian Schulze-Hagen}
\author[4]{Michael Ho}
\author[4]{Christoph Haarburger}
\author[5,6,7]{Fabian Kiessling}
\author[3,7]{Christiane Kuhl}
\author[1,6,7,*, $\dagger$]{Volkmar Schulz}
\author[3,4, *, $\dagger$]{Daniel Truhn}
\date{}
\affil[1]{\small Physics of Molecular Imaging Systems, Experimental Molecular Imaging, RWTH Aachen University, Germany}
\affil[2]{\small Department of Diagnostic and Interventional Radiology, University Hospital Düsseldorf, Germany}
\affil[3]{\small Department of Diagnostic and Interventional Radiology, University Hospital Aachen, Germany}
\affil[4]{\small ARISTRA GmbH, Berlin, Germany}
\affil[5]{\small The Institute for Experimental Molecular Imaging, RWTH Aachen University, Germany}
\affil[6]{\small Fraunhofer Institute for Digital Medicine MEVIS, Bremen, Germany}
\affil[7]{\small Comprehensive Diagnostic Center Aachen (CDCA), University Hospital RWTH Aachen, Aachen, Germany}
\affil[$\dagger$]{\small Both authors contribute equally}
\affil[*]{Correspondence should be addressed to V.S (schulz@pmi.rwth-aachen.de) and D.T. (dtruhn@ukaachen.de)}

\begin{document}
\maketitle
\begin{abstract}

\textbf{
	Unmasking the decision making process of machine learning models is essential for implementing diagnostic support systems in clinical practice. 
	Here, we demonstrate that adversarially trained models can significantly enhance the usability of pathology detection as compared to their standard counterparts.
	We let six experienced radiologists rate the interpretability of saliency maps in datasets of X-rays, computed tomography, and magnetic resonance imaging scans.
	Significant improvements were found for our adversarial models, which could be further improved by the application of dual-batch normalization. 
	Contrary to previous research on adversarially trained models, we found that accuracy of such models was equal to standard models, when sufficiently large datasets and dual batch norm training were used.
	To ensure transferability, we additionally validated our results on an external test set of 22,433 X-rays.
	These findings elucidate that different paths for adversarial and real images are needed during training to achieve state of the art results with superior clinical interpretability.
}
\end{abstract}

\section*{INTRODUCTION}
Computer vision (CV) in medical imaging has been a focus of radiological research in recent years.
It is likely that CV methods will soon be used as adjunct tools by radiologists:
Computer-aided diagnosis can help to speed up the diagnostic process by guiding radiologists to findings worth looking at and maximize diagnostic accuracy by reducing subjectivity \cite{lundberg2018explainable, he2020integrating, truhn2019radiomic, rajpurkar2017chexnet, rajpurkar2018deep}.
Prominent examples are deep convolutional neural networks (CNN), which had their breakthrough when more conventional computer vision algorithms were far surpassed by residual neural networks in 2015 \cite{he2016deep}.
Similar developments have taken place in medicine, where CNNs performed comparable to experts in lung cancer diagnosis \cite{ardila2019end, coudray2018classification, levine2019rise}, retinal disease detection \cite{de2018clinically, tsao2018predicting, arcadu2019deep}, and skin lesion classification \cite{esteva2017dermatologist, tschandl2020human, liu2020deep}.

However, state-of-the-art deep learning models behave like a black box in that their decision process remains obscure and inexplainable to human readers \cite{rudin2019stop}.
This gives rise to problems: First, deep learning models trained in a standard fashion are vulnerable when facing attacks from adversaries.  
An attacker might introduce a subtle change into the image - such as changing a single pixel \cite{su2019one} - and manipulate the output of the model towards a desired direction, e.g., non-presence of disease when in reality the disease is present. 
Recent studies suggest that adversarial attacks are capable of manipulating predictions of deep learning models across diverse clinical domains \cite{finlayson2019adversarial, han2020deep}. 
The second problem is associated with the so-called ‘Clever Hans’-type decision strategy \cite{lapuschkin2019unmasking} where a model focusses its attention towards irrelevant details such as a hospital department's tag present in the X-ray when assessing the severity of diseases.
Thus, patients with X-rays from the intensive care unit might seemingly correctly get classified as pathological even though the model does nothing else than to scan for that specific tag.
For the clinician, it is thus important to understand, why a model arrives at a certain conclusion and if that reasoning aligns with human reasoning:
A reliable and robust mechanism to explain the model's reasoning could support the acceptance of deep learning models in clinical routine \cite{price2018big}.
Numerous efforts have been devoted to understanding such decision making processes and to find an agreement between data-driven features and human expertise \cite{zhang2019pathologist, tang2019interpretable, lee2019explainable, selvaraju2017grad, hosny2018deep}.

Our hypothesis is that robustly trained models, i.e. models that are to a certain extent immune to adversarial attacks, are not always at odds with accuracy as previously assumed.
We examine, if this conflict can be resolved when employing specific training paradigms and having sufficient amounts of radiological data available.
Moreover, we hypothesize that training the model in this fashion allows for the model's reasoning to be more closely aligned with clinical expectations than when the model is trained in a conventional fashion.
We test these hypotheses in dedicated experiments on X-rays, computer tomography images and magnetic resonance images and involve six radiologists who rate the clinical validity of our results.

\section*{RESULTS}
\subsection*{Robustness and Model Performance Trade-off}

\begin{figure}[h!]
	\centering
	\scalebox{1.0}{
		\includegraphics[trim=0 0 150 0, clip, width=\textwidth]
		{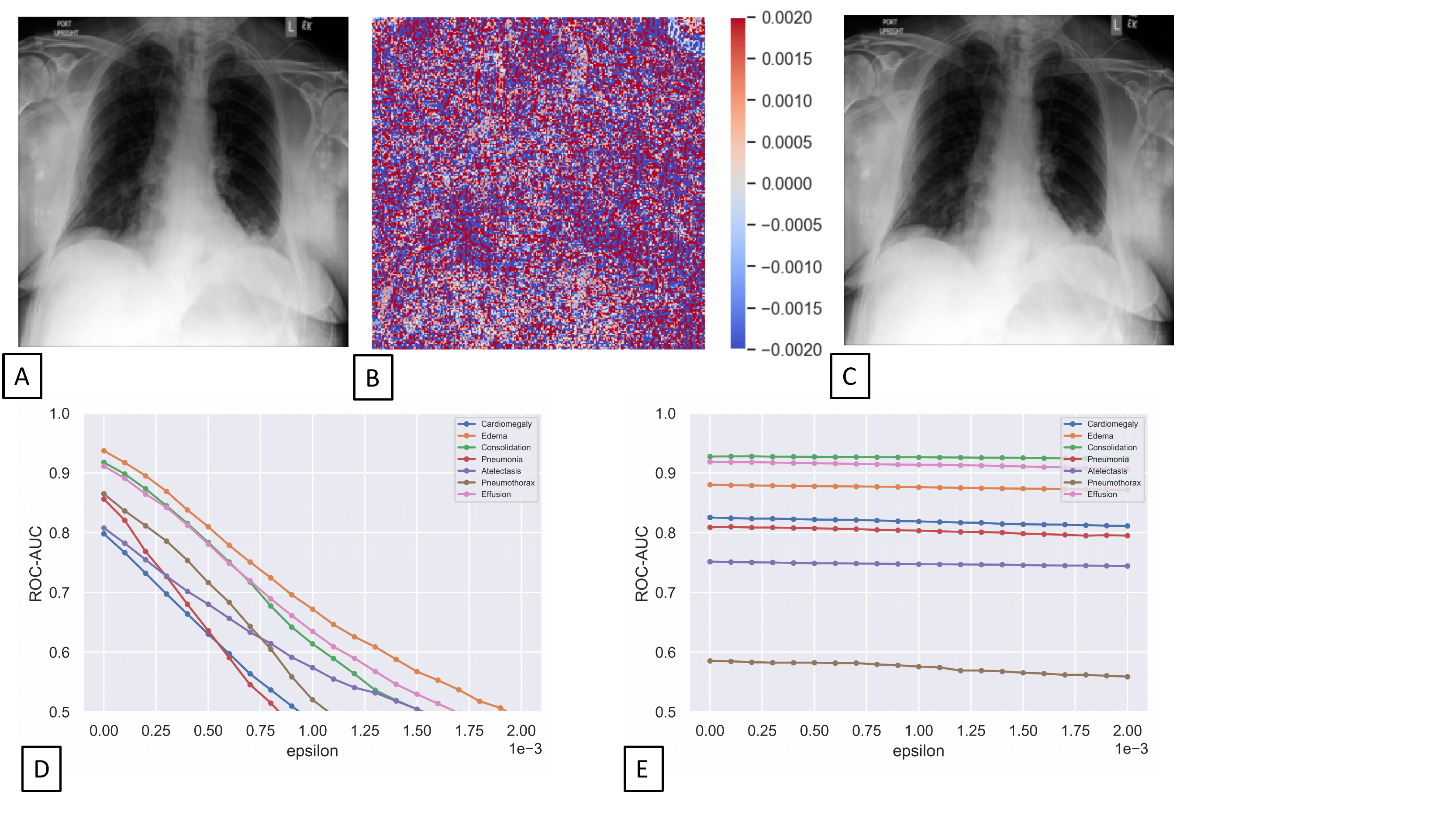}}
	\caption{\label{fig:auc_epsilon}\textbf{Adversarially trained models are robust against adversarial attacks.} 
		Adversarial perturbations with increasing strength ($\epsilon$) were generated via an iterative fast gradient sign method (FGSM).  
		To demonstrate the impact of adversarial attacks on state-of-the-art classifiers, we trained a ResNet-50 models with a large chest X-ray dataset (CheXpert) containing nearly 200,000 X-rays.  
		(A) Original unmanipulated chest radiograph.
		(B) Adversarial noise with $\epsilon$ = 0.002.
		(C) Manipulated chest radiograph (original radiograph + noise), i.e. adversarial example. 
		(D) The standard model was easily misled by small adversarial perturbations (B) that are not perceptible to the human eye (C) and accuracy in classifying the disease dropped drastically when allowing more pronounced perturbations. 
		(E) Only a limited amount of performance degradation was observed when applying adversarial attacks on the model trained adversarially ($\epsilon$ during training was set to 0.005).}
\end{figure}

Supervised models are vulnerable to adversarial attacks in which an adversary subtly changes the input to the model and thereby manipulates the prediction of that model.
Instead of optimizing the parameters $\theta$, e.g., the weights of the neurons, of a model $h_{\theta}$ towards the minimum of the loss function $\mathcal{L}$, 
\begin{equation}
	\label{equ:loss}
	\underset{\theta}{\text{minimize}} \:\underset{(x, y) \sim D}{\mathbb{E}} \mathcal{L}(h_{\theta}(x), y),
\end{equation} 
one is able to generate adversarial examples $(x+\delta)$ by solving the optimization problem
\begin{equation}
	\label{equ:attack}
	\underset{\delta\in\Delta}{\text{maximize}}\: \mathcal{L}(h_{\theta}(x+\delta), y),
\end{equation} 
where $(x, y)$ is an input-label pair in the dataset D, $\delta$ is the applied adversarial perturbation, and $\Delta$ is an allowable set of perturbations. 
In practice, adversarial examples will always be designed to be as unconspicious as possible to the human eye.
One commonly define the allowed perturbations set $\Delta$ to be a hypersphere ball around any data $x$ with a constrained norm ($l_{\infty} \leq \epsilon$) \cite{goodfellow2014explaining}.
To select the best model in the task of pathology detection, we quantified the model's performance via the area under the receiver operating characteristic curve (ROC-AUC). 
In accordance with previous research, we found that adversarial perturbations can easily influence conventionally trained models when applied to disease detection in thoracic X-rays, see Fig \ref{fig:auc_epsilon}. D:
the standard classifier was significantly biased even by a small amount of adversarial perturbation ($\epsilon \leq 0.001$).

In a second experiment, we made the models robust to adversarial attacks by employing an approach proposed by Madry et al. \cite{madry2017towards}.
In this approach, we minimized the expected adversarial loss via performing gradient descend on adversarial samples - effectively presenting the model with adversarial examples during training:
\begin{equation}
	\underset{\theta}{\text{minimize}} \underset{(x, y) \sim D}{\mathbb{E}}[\underset{\delta \in \Delta}{\text{maximize}} \:\mathcal{L}(h_{\theta}(x + \delta), y)].
\end{equation}

\begin{figure}[h!]
	\centering
	\scalebox{1.0}{
		\includegraphics[trim=40 600 300 0, clip, width=\textwidth]
		{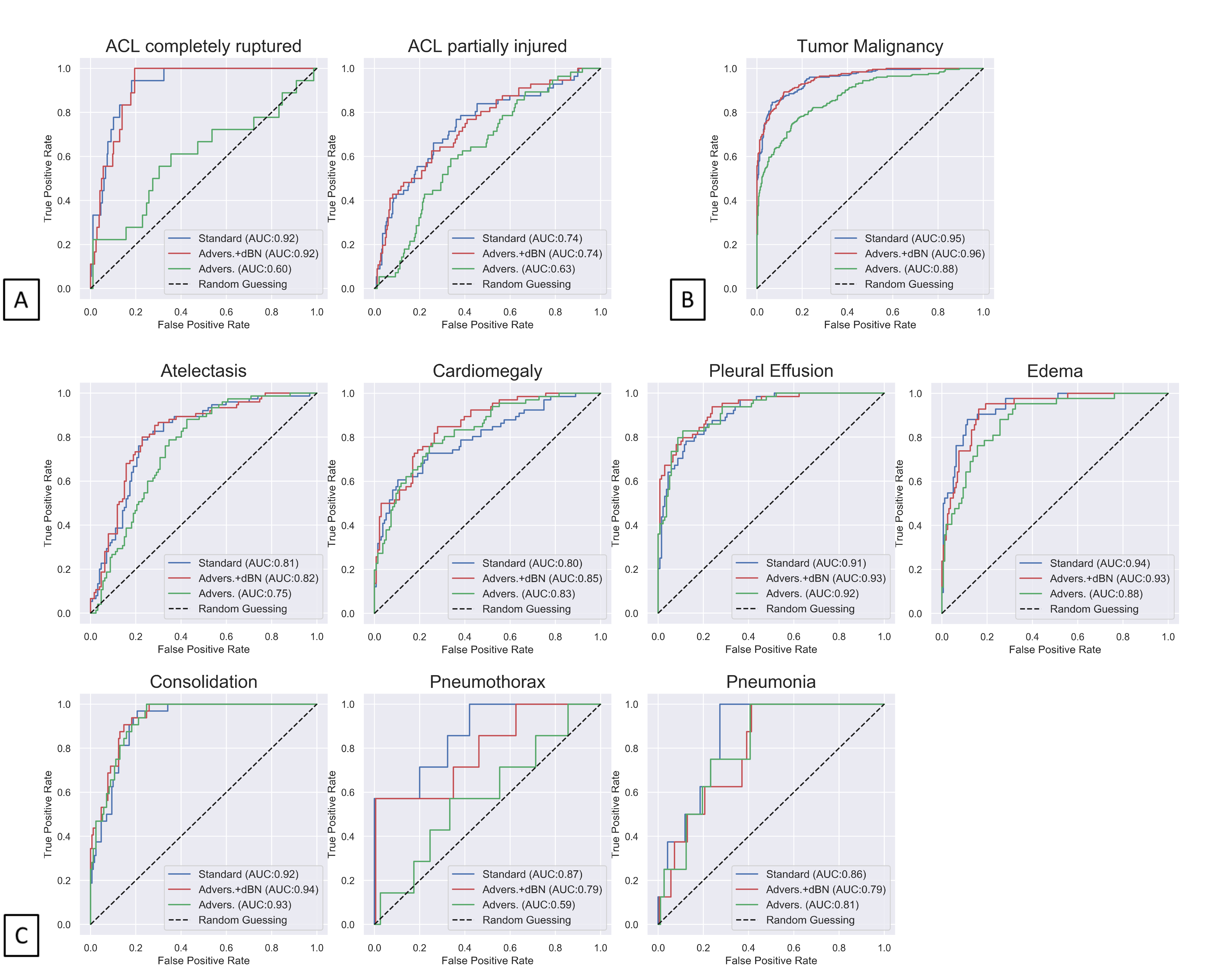}}
	\caption{\label{fig:aucs}\textbf{The use of dual batch norm boosts the classification performance of neural networks.}
		Three models were compared: blue: neural network without adversarial training, green: neural network with adversarial training and red: neural network with adversarial training employing dual batch norms.
		The models' performances were tested on three distinct datasets:
		(A) Rijeka knee magnetic resonance imaging (MRI) dataset. 
		(B) Luna16 dataset containing computed tomography (CT) slices of malignant tumors and (C) CheXPert thoracic X-ray dataset. 
		In line with previous reports \cite{tsipras2018robustness} we found that robustness and good performance appear to be incompatible when data is limited. 
		In both experiments A and B, the AUC of naively adversarially trained models (green) dropped significantly as compared to models trained in a standard fashion (blue).
		However that performance gap was reduced when the models were trained on a large dataset of 191,027 radiographs C.
		Adversarially trained models performed best when employing dual batch norm (red), no significant difference in performance to the naively trained models were found.
		As reflected by the red curves, the performance of robust models was boosted across different datasets when dual batch norm training was employed (A-C).
		}
\end{figure}

Fig. \ref{fig:auc_epsilon} visualizes our result, that an adversarially trained (robust) classifier (Fig. \ref{fig:auc_epsilon}E) was less sensitive to adversarial perturbations than its counterpart that was trained in a standard fashion (Fig. \ref{fig:auc_epsilon}D). 
Nevertheless, as other groups have previously shown, that robust models appear to be less accurate than standard models \cite{tsipras2018robustness, zhang2019theoretically, su2018robustness}.
We tested these results in the context of medical datasets by performing adversarial training on the Luna16 \cite{setio2017validation}, kneeMRI \cite{vstajduhar2017semi}, and CheXpert \cite{irvin2019chexpert} datasets (shown in Fig. \ref{fig:aucs}).  
We found that robust models were indeed less accurate when trained on limited datasets, see blue and green curves in Fig. \ref{fig:aucs} A and B.  
However, we found that the performance gap between standard and robust models was less pronounced when sufficient amounts of data were available, see Fig. \ref{fig:aucs} C.

\subsection*{Achieving both Robustness and Accuracy}
Adversarial examples can also be viewed as a form of data augmentation guided by gradient backpropagation \cite{tsipras2018robustness, miyato2018virtual}.
However, adversarial training faces problems due to differing statistics between the original data and the adversarially perturbed data \cite{xie2019intriguing}.
To optimize the gradient flow of adversarial training, we used separate batch normalization layers for the aim of reparametrizing original ($x$) and adversarially perturbed ($x^{\ast}$) batches \cite{xie2019intriguing, santurkar2018does}. 
Giving the network the freedom to choose the reparameterization parameter $\gamma$ independently for adversarial and original batches resulted in differing distributions, see Fig. \ref{fig:gamma}.
More importantly, the employment of separate normalization layers removed the performance gap towards naively trained models:
as demonstrated by the red AUC in Fig. \ref{fig:aucs}, no performance difference towards naively trained models was found, even when the size of the training set was comparatively small.
A more detailed summary of performance metrics and confidence intervals can be found in \cref{Table: luna,Table: knee,Table: chexpert}.
No significant differences in ROC-AUC, sensitivity, and specificity were found when comparing the naively trained non-robust model to the adversarially trained model with dual batch norm training.

\subsection*{Generalization of Robust Models}
\begin{figure}[h!]
	\centering
	\scalebox{1.0}{
		\includegraphics[trim=30 220 320 200, clip, width=\textwidth]
		{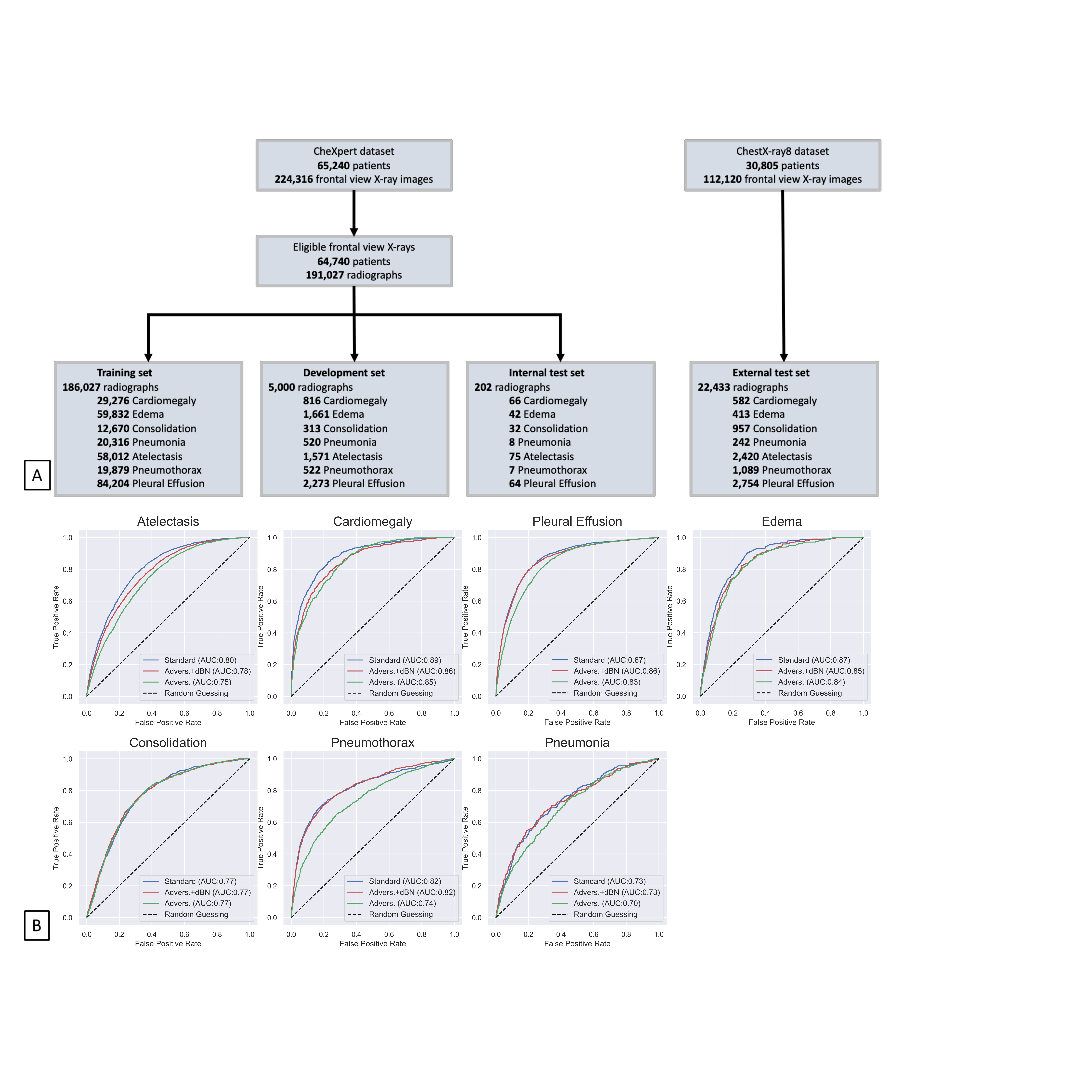}}
	\caption{\label{fig:extern}\textbf{Validation of models on an external dataset (ChestX-ray8).}
		(A) Schematic of the data selection process.
		(B) AUC of the standard model (blue) and the adversarially trained models with (red) and without (green) dual batch norms on an independent test set of 22,433 radiographs from the ChestX-ray8 dataset.
		Dual batch norm training resulted in better AUC, closely matching the performance of the standard model.
	}
\end{figure}

In general, overfitting of machine learning models can be a problem.
To explore whether this was the case with our robust training, we used the external ChestX-ray8 dataset as a test dataset.
A comparison of the distribution of ground-truth labels of both datasets is listed in Table \ref{Table: xrays_meta}.
Fig. \ref{fig:extern} shows a comparison of the standard and robust models solely trained on the CheXpert dataset.
Models were validated on the external ChestX-ray8 dataset that contained 22,433 radiographs and had never before been presented to the models \cite{wang2017chestx}.  
As before, the robust model employing dual batch norms outperformed the conventional adversarially trained model and performed comparably to the standard model.

\subsection*{Adversarial Training Selects Clinically Meaningful Features}
\begin{figure}[h!]
	\centering
	\scalebox{1.0}{
		\includegraphics[trim=20 460 150 50, clip, width=\textwidth]
		{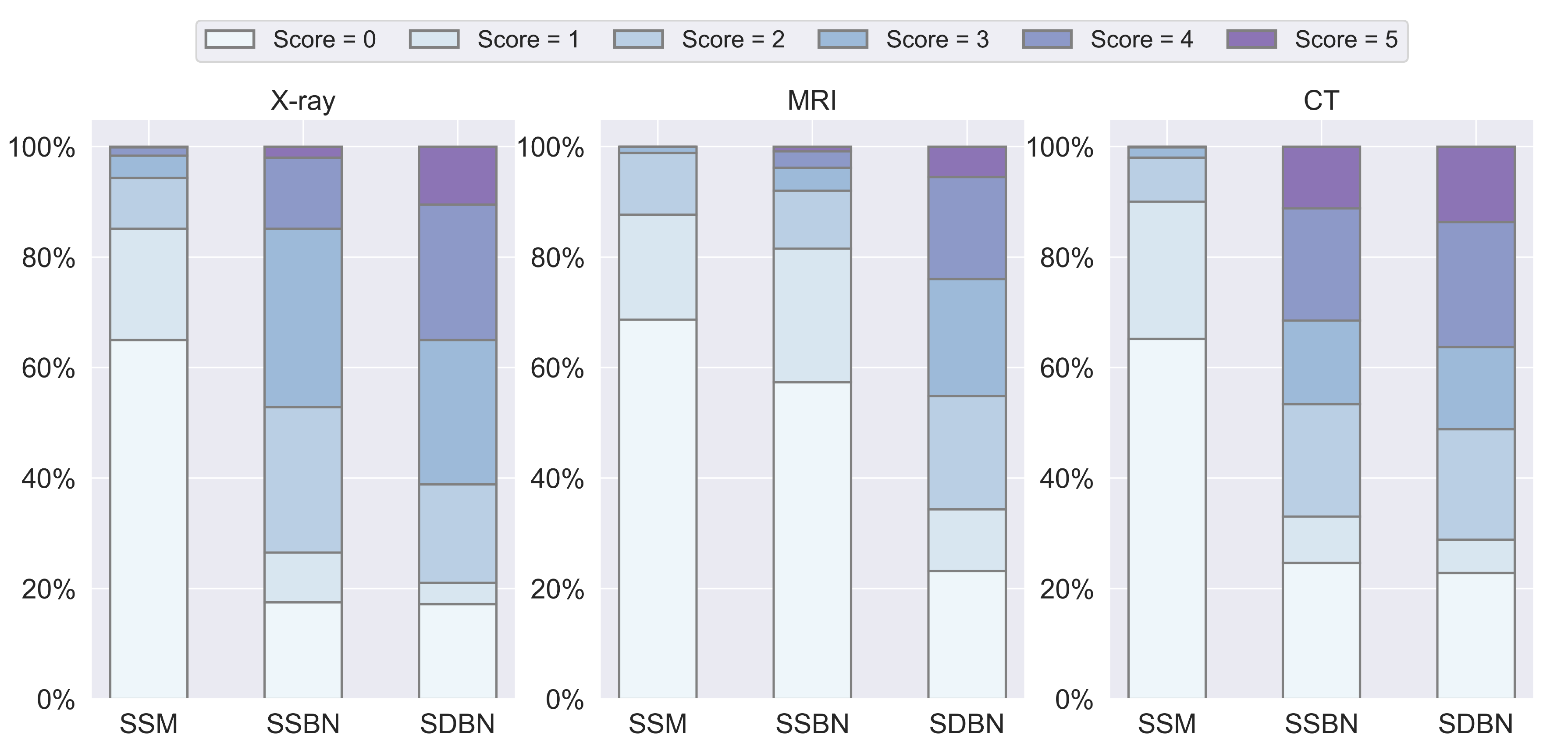}}
	\caption{\label{fig:rater_dist}\textbf{Adversarially trained neural network with dual batch norm yields clinically interpretable saliency maps.} 
		Figure shows the assessment of diagnostic relevance in percentage for SSM, SSBN, and SDBN models as evaluated independently by six radiologists. 
		Each color bar reveals percentage of gradient saliencies with same rating score.  
		}
\end{figure}

\begin{figure}[h!]
	\centering
	\scalebox{1.0}{
		\includegraphics[trim=350 1400 700 10, clip, width=\textwidth]
		{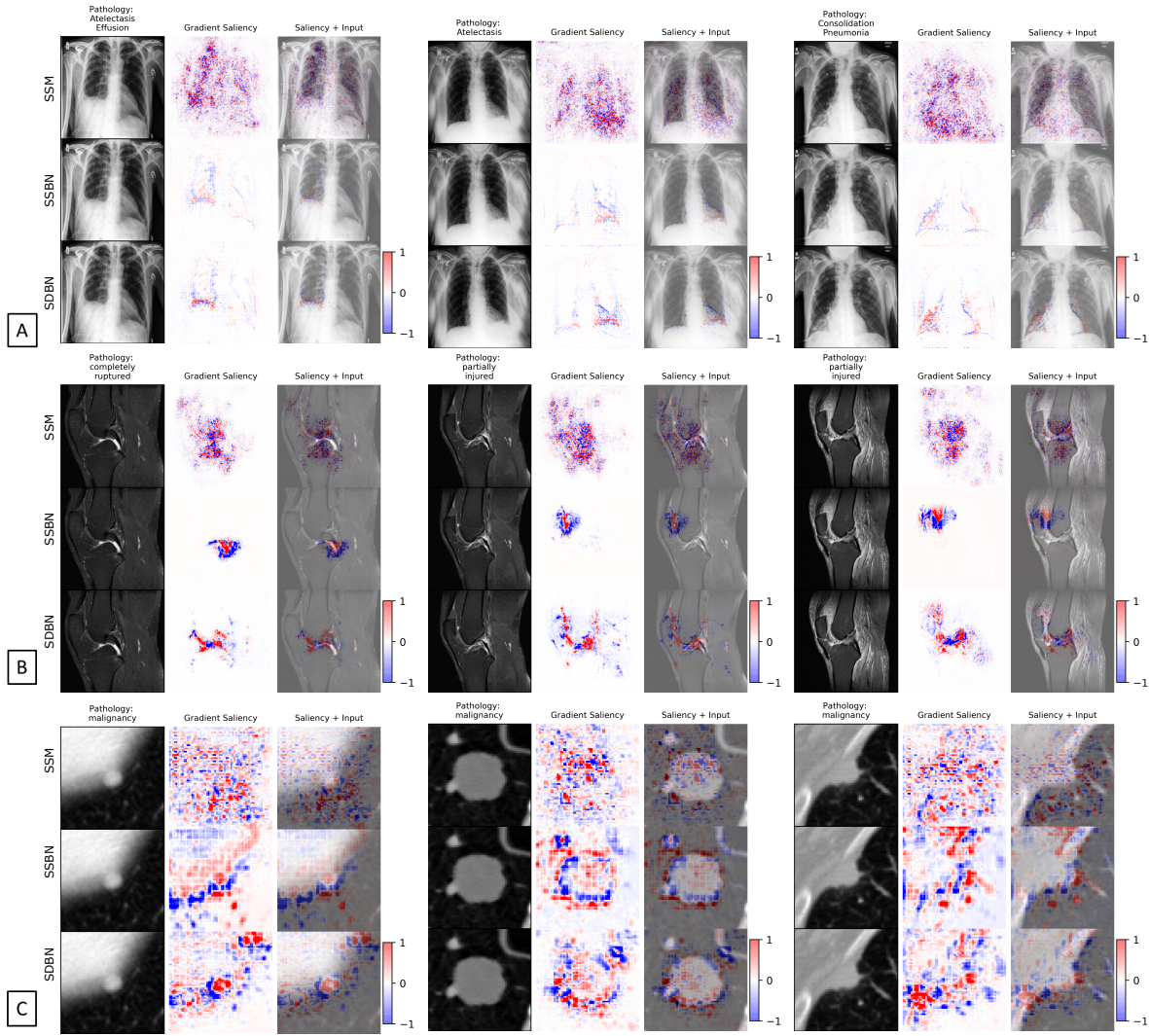}}
	\caption{\label{fig:saliency}\textbf{Saliency maps can help in guiding specialists to the correct diagnosis.}
		Loss gradients were plotted with respect to their input pixels for the X-ray (A), MRI (B), and CT (C) datasets.
		No extra preprocessing steps were applied to the loss gradients. 
		Here, blue and red colored pixels denote negative and positive gradients individually. 
		For comparison, saliency maps of standard and robust models are presented next to the radiological input. 
		In all three datasets, the SDBN pointed more accurately and more distinctively to the pathology than the SSBN, while the SSM was almost completely uncorrelated and almost noise-like.
		It can also be observed that the saliency maps created based on neural networks that have been trained on a very large dataset of hundreds of thousands of images (A) were more precise in pointing to the pathology, than those trained on datasets containing fewer examples (B and C).  
		}
\end{figure}

Machine learning is often seen as a black box algorithm \cite{price2018big} in the sense that the way in which a machine learning model comes to its result is not transparent.
This hinders the acceptance of such algorithms, as humans put more trust in conclusions which are comprehensible.
Furthermore, such information might be helpful in pointing the radiologist to the suspicious finding, e.g., a tumor, in the radiological examination.
To visualize the parts of the image which are decisive for the algorithm's conclusions, we highlighted the input pixels that contribute most to the final model prediction via gradient backpropagation.
These pixel maps are then displayed in a color map and are generally referred to as saliency maps. 
As previously reported by Tsipras et al. \cite{tsipras2018robustness} saliency maps of robust models were found to be both sparse and well aligned with human expertise, see Fig. \ref{fig:saliency}.
In this study, it was found that the saliency maps of the adversarially trained neural network with dual batch norm (SDBN) agreed significantly better with human expertise than both the adversarially trained models with a single batch norm (SSBN) and those of the standard model (SSM):
Six radiologists were given the task to rate the meaningfulness of the saliency maps in guiding the radiologist to the correct pathology on a scale from 0 (no correlation between pathology/ies and hot spot(s) on saliency map) to 5 (clear and unambiguous correlation between pathology/ies and hot spot(s) on saliency map).
Across all modalities - i.e. X-ray, MRI, and CT - all readers consistently found that SDBN were more meaningful than SSBN which in turn were more meaningful than the SSM, see table \ref{table:meanRatings}.
These results were most significant for X-rays, followed by MRIs and CTs, most likely due to the number of training samples:
for the training of the classifiers, about 200,000 X-rays were used, while only 917 MRIs and 888 CT scans were available respectively.
The results of individual readers as well as the pooled ratings are given in table \ref{table:individualRatings}.
Fig. \ref{fig:rater_dist} visualizes the distribution of all ratings for X-rays, MRIs, and CTs respectively.
In all modalities, the highest ratings were achieved for SDBN, with significant proportions of the maps reaching ratings as high as four and five, while only negligible fractions reached higher ratings for the standard SSM.


\begin{table}[h!]
	\centering
	\captionof{table}{Mean ratings of radiologists in guiding the radiologist to the correct pathology on a scale from 0 (useless) to 5 (saliency map points clearly and unambiguously to the correct pathology). In total 100 images were rated by 6 radiologists for each dataset}
	\begin{tabular}{ccccc}
	Dataset & Score for SSM & Score for SSBN & Score for SDBN & Friedman test \\ \hline
	\rowcolor[HTML]{EFEFEF}
	X-ray & $0.57 \pm 0.94$ & $2.20 \pm 1.33$ & $2.69 \pm 1.56$ & $p < 0.001$ \\
	MRI & $0.49 \pm 0.74$ & $0.74 \pm 1.09$ & $2.17 \pm 1.57$ & $p < 0.001$ \\
	\rowcolor[HTML]{EFEFEF}
	CT  & $0.47 \pm 0.73$ & $2.32 \pm 1.72$ & $2.50 \pm 1.74$ & $p < 0.001$ \\ \hline
	\end{tabular}
	\label{table:meanRatings}
\end{table}

More precisely, a robust model, as shown in red in Fig. \ref{fig:saliency} A, was able to detect common thoracic diseases such as cardiomegaly, atelectasis, and pneumonia based on the organ shape and the lung opacity.
While the SDBN pointed more clearly to the areas of interest that were decisive for diagnosis, the SSN had less focus on these areas and the SSM was almost completely uncorrelated to these areas and useless in guiding the radiologist to the correct conclusions.
Similarly, for knee magnetic resonance images (Fig. \ref{fig:saliency} B), the SDBN showed a more direct correlation with the pathology than the SSBN, which more often pointed to accompanying, but more unspecific phenomena such as joint effusion.
Again, the SSM was almost completely uncorrelated to the imaging pattern of the disease.
Finally, for the intrapulmonary malignancies in CT slices shown in Fig. \ref{fig:saliency} C, the finding that SDBN, SSBN, and SSM were useful in descending order was again confirmed with the borders of the malignancy being emphasized more pronounced in the SDBN.
These examples also illustrate that large data is needed to train a network that generates meaningful maps:
while the examples of Fig. \ref{fig:saliency} A showed a good correlation between the disease pattern and the saliency maps - at least for SDBN and SSBN - the correlation was less pronounced for the dataset of knee MRI and CT slices (Fig. \ref{fig:saliency} B and C) due to the much smaller size of those datasets.


The above finding suggests that the reparameterization (learnable parameters $\gamma$ and $\beta$, supplement section \nameref{Section: repara}) offered by batch norms affect the representation learned by deep neural networks. 

\subsection*{Batch Normalization Influences Learned Representations}

\begin{figure}[h!]
	\centering
	\scalebox{1.0}{
		\includegraphics[trim=0 0 330 0, clip, width=\textwidth]
		{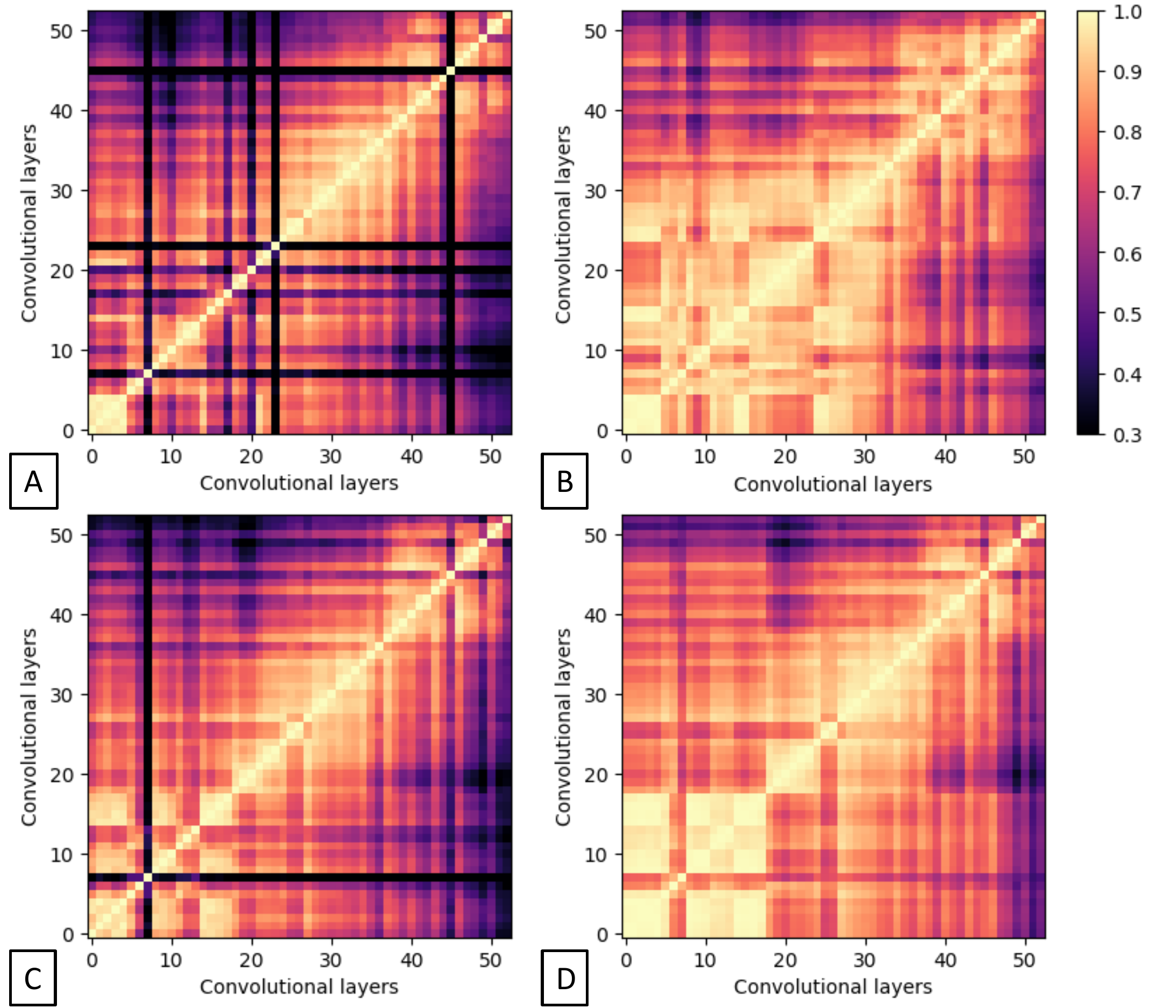}}
	\caption{\label{fig:representations}\textbf{Linear  centered kernel alignment (CKA) reveals representations are influenced by batch norms.}
	To explore the learned hidden representations, the linear CKA between convolutional layers of the models was computed on the CheXpert test set:
	a model trained with a single batch norm in a conventional setting with real examples (A), a model trained with a single batch norm with real and adversarial examples (B), and a model trained with a dual batch norm with real and adversarial examples when the respective CKA was evaluated separately with the batch norm used for real (C) and adversarial (D) examples.
	It should be noted that the observed grid pattern in A was due to the residual connections in the ResNet architecture \cite{kornblith2019similarity}.
	When employing adversarial training with a single batch norm, layers of the network seem to get more similar to each other, as visualized by the block-like structure arising from the high degree of similar neural activations in Fig. B.
	This indicates, that the neural network loses complexity due to adversarial training which might contribute to a loss in performance.
	When employing a dual batch norm for original and adversarial examples respectively, the complexity of the network seems to be preserved (note the similarity between A and C), when presented real examples using the first batch norm, while simultaneously robustness to adversarial examples arises due to the same changes when employing the second batch norm (D) that the network from B underwent (note the similarity between B and D). 
	}
\end{figure}

In the setting of adversarial training with dual batch norms, original and perturbed batches are decoupled via passing them through separate batch norm layers. 
To understand how the use of these two batch norms influenced the features learned by the robust models, we quantified the similarity between the layers within the respective deep neural network by using the linear centered kernel alignment (Linear CKA) method \cite{kornblith2019similarity}.
In Fig. \ref{fig:representations} A, we visualized the typical representation learned by a model in a standard training setting, with only one batch norm.
We found that a certain degree of correlation between succeeding layers was present.
However, long-range correlations - i.e. correlations between layers that were far apart - tended to be relatively weak, indicating, that the information that was passed on in the network gets continuously processed.
The situation was different however for the same network architecture when adversarial training with only one batch norm was used, see Fig. \ref{fig:representations} B:
Long ranging correlations resulted in a block like structure and the first 35 layers (about 65 \% of the network) seemed to carry approximately the same information.
Such a high similarity of learned representations may be part of the reason for the performance degradation of robust models trained via vanilla adversarial training \cite{tsipras2018robustness, kornblith2019similarity}: it seems that the networks may not be able to encompass the full complexity of the dataset after adversarial training.
The network seems to effectively reduce to a simpler - less deep network since neighboring layers contained similar activations.
Using a dual set of batch norms for the original image samples and the adversarial image samples seemed to preserve the complexity of the network when fed with the original samples (Fig. \ref{fig:representations} C), while at the same time providing the same transition as in Fig. \ref{fig:representations} B for the adversarial samples as indicated by the similarity between the linear CKA maps of Fig. \ref{fig:representations} B and D.

\section*{DISCUSSION}
The purpose of this study was to investigate the applicability and potential advantages of adversarially robust models in the field of medical imaging.
A limitation of deploying such models in clinics is a potential performance degradation as compared to conventionally trained models that has been found by other research groups \cite{tsipras2018robustness}.
In our experiments shown in Fig. \ref{fig:aucs}, however, we found that this effect appears almost negligible when training the models on large image data sets (\cref{Table: luna,Table: knee,Table: chexpert}) and when applying dual-batch norms, i.e., no significant difference in the AUC was found between the standard model and the adversarially trained model with dual batch norms.
Furthermore, we have validated that robust models can generalize well on external datasets by employing 22,433 X-rays from the ChestX-ray8 dataset, that had not been part of the training process and originated from a different institution.
Most likely, the reason that other groups had found significant differences between the performance of standard models and adversarially trained models is the use of a single batch norm in adversarial training:
we consistently found in all our experiments, that to achieve the best results in adversarial training, it was necessary to employ separate batch norms for real and adversarial examples.
By using dual batch norms, our adversarially trained model achieved state-of-the-art results on thoracic pathology detection \cite{irvin2019chexpert}.

Not only are adversarially trained models less vulnerable to adversarial attacks (see Fig. \ref{fig:auc_epsilon}), but saliency maps generated by adversarially trained models provide significantly more information to the clinicians than those generated by standard models and may help to guide them to the right diagnosis.
This can also boost the acceptance of deep learning models in clinical routine.
Deep learning models are often regarded as a black box and not much trust is put into their opaque decision-making process by clinicians.
By providing the clinician with a meaningful saliency map as generated by adversarially trained deep neural networks with dual batch norms, the decision of the neural network can be made more transparent resulting in a better acceptance by experts.

We further investigated a potential reason for the better performance of adversarially trained models with dual batch norms:
while conventional adversarial training seems to reduce the complexity of the neural network as indicated by the increased long-range correlations of the linear CKA between layers of the network (see Fig. \ref{fig:representations}), the use of dual batch norms preserves complexity levels of the networks when feeding in real examples, while simultaneously accommodating for the increased robustness to adversarial examples. 

This study is limited by the use of neural networks using two-dimensional inputs.
Medical data in CT, MRI, and positron emission tomography (PET) is inherently volumetric (3D) or even volumetric plus time (4D) and it might be expected that models encompassing such higher dimensional inputs (instead of a series of two-dimensional slices) can improve upon their performance. 
More research is needed to train a model with high dimensional inputs as adversarial training commonly becomes more difficult in a high-dimensional feature space. 
If higher-dimensional models become more widespread and applicable, future studies should try to reproduce our results in such models.

In conclusion, we demonstrated, that adversarially trained models with a dual batch norm are not only equivalent to standard models in terms of diagnostic performance but offer additional advantages in conveying their reasoning through the use of clinically useful saliency maps and being more robust to adversarial attacks.
We encourage fellow research groups to employ adversarially trained neural networks in their applications and hope that this will not only lead to more robust and better results in terms of diagnostic performance, but also increased acceptance of such algorithms in clinical practice.


\section*{MATERIALS AND METHODS}
\subsection*{Study Datasets}
A total number of four medical imaging datasets are used in this study: the CheXpert dataset, which has been released by Irvin et al. in January 2019 and contains 224,316 chest radiographs of 65,240 patients \cite{irvin2019chexpert}. 
Only 191,027 frontal radiographs are downloadable for model training.
To clean up CheXpert labels, we assigned pathology labels not mentioned to 0.0.
According to the labeling performance comparison \cite{irvin2019chexpert}, the uncertainty labels (U) were assigned to 1.0, except for the consolidation class (to 0.0). 
For testing, we compare the performance of the trained models on the official validation set of 202 scans on which the concurrence of diagnosis from three radiologists serves as ground truth \cite{irvin2019chexpert}.
Another X-ray dataset used in this study is the ChestX-ray8 dataset released by the National Institutes of Health (NIH) in 2017, containing 112,120 frontal radiographs of 30,805 unique patients \cite{wang2017chestx}. 
We randomly select 20\% out of 112,120 radiographs, i.e., 6,187 patients and 22.433 radiographs, to form an external test set. 
The seven overlapping labels between the CheXpert and ChestX-ray8 datasets are listed in table \ref{Table: xrays_meta}.

The MRI dataset used in this study is the kneeMRI dataset, which has been released by Štajduhar et al. \cite{vstajduhar2017semi} in 2017 and contains 917 sagittal proton-density weighted knee scans from Clinical Hospital Centre Rijeka, Croatia.
Three degrees of anterior cruciate ligament (ACL) injuries were recorded by radiologists in the Rijeka dataset: namely non-injured (692 scans), partially injured (172 scans), and completely ruptured (55 scans).
Bounding box annotations of the region of interest (ROI) slices responsible for ACL tear diagnosis were also provided along with labels.
According to ROI labels, we extracted a total of 3,081 diagnostic relevant MRI slices and randomly split them into 80\% training, 10\% development, and 10\% testing. 
Lastly, we investigated the applicability of adversarial training on CT data with the Luna16 dataset consisting of 888 CT scans with lung cancer ROI annotations \cite{setio2017validation}.
In total, a number of 6,691 lung cancer patches were extracted and randomly split into 80\% training, 10\% development, and 10\% testing.

\subsection*{PGD Attack and Adversarial Training}
Following equation \ref{equ:loss}, we let the model parameter be denoted as $\theta$, model loss as $\mathcal{L}$ and training input\&label as (x, y).
The projected gradient descent (PGD) method repeatedly adjusts the model's inputs x in the direction of maximizing the loss function, i.e., $sign(\nabla_{x} \mathcal{L}(\theta, x, y))$.
To safeguard models against adversarial threats, we trained our models against a PGD adversary via both vanilla and dual batch norm adversarial training. 
The details of the adversarial training procedure with separate batch norms are depicted in algorithm \ref{algo:advtrain}.

\begin{algorithm}[h]
	\caption{\textbf{Adversarial Training with Separate $\textbf{BN}_{\textbf{std}}$ and $\textbf{BN}_{\textbf{adv}}$}. We use default values of $\epsilon$ = 0.005, $\alpha$ = 0.0025, $k$ = 10, and $b$ = 64}\label{algo:advtrain}
	\begin{algorithmic}[1]
		\State \textbf{Require:} Dataset D;  A model h with its current parameter $\theta$ and loss function $\mathcal{L}$;
		batch norm layers for standard inputs $BN_{std}$; batch norm layers for adversarial inputs $BN_{adv}$; the batch size $b$; learning rate $\eta$.
		\State \textbf{Require:} $l_{\infty}$ boundary $\epsilon$; stepsize $\alpha$; number of attack iterations $k$.
		\State Sample a batch of inputs $\{x^{(j)}\}_{j=1}^b \sim D$ and class labels $\{y^{(j)}\}_{j=1}^b \sim D$.
		\For {$j = 1, ..., b$}
		\State Sample input $x \sim \{x\}^b$ and label $y \sim \{x\}^b$
		\State $x_0^{\ast} = x$
		\For {$t = 1, ..., k$}		
		\State Input $x_{t-1}^{\ast}$ to model $h_{\theta}(x_{t-1}^{\ast})$
		\State $x_t^{\ast} \gets x_{t-1}^{\ast} + \alpha \text{sign}(\nabla_{x}(\theta, x_{t-1}^{\ast}, y))$
		\State $x_t^{\ast} \gets \text{clip}_{x, \epsilon}(x_t^{\ast})$
		\EndFor
		\State \textbf{end for}
		\EndFor
		\State \textbf{end for} 
		\State $\{x^{\ast}\}^b$ = $\{x_k^{\ast}\}^b$
		\State $\mathcal{L}_{std} = \mathcal{L}(\theta, BN_{std}, \{x\}^b, \{y\}^b)$; $\mathcal{L}_{adv} = \mathcal{L}(\theta, BN_{adv}, \{x^{\ast}\}^b, \{y\}^b)$
		\State $\theta \gets \theta + \eta \nabla_{\theta} (\mathcal{L}_{std}^j + \mathcal{L}_{adv}^j)$ 
	\end{algorithmic}
\end{algorithm}

\subsection*{Model Architecture and Training}
We used ResNet-50 architecture for our experiments in this study. 
For all classification tasks, an Adam optimizer with default $\beta_1$ = 0.9, $\beta_2$ = 0.99, and $\epsilon$ = 1e-8 \cite{kingma2014adam} was used to optimize the loss.
In a total number of 300 training epochs, We decayed the initial learning rate 0.01 by a factor of 10 once the number of epochs reached 100 epochs.
All classifier models utilized development-based early stopping with sigmoid binary cross-entropy loss as the criterion.

Medical images from CheXpert, ChestX-ray8, and kneeMRI were scaled to a fixed resolution of 256$\times$256 pixels whereas tumor patches extracted from Luna16 ROI slices were scaled to 64$\times$64 pixels.
During training, random color transformations such as adjusting contrast, brightness, saturation, and hue factor were applied to each training image. 
Additionally, we also performed spatial affine and random cropping augmentations before normalizing each input to the range of 0 to 1.  

All computations were performed on a GPU cluster equipped with two Intel Xeon(R) Silver 4208 processor (Intel, Santa Clara, Calif) and three Nvidia Titan RTX 24 GB GPUs (Nvidia, Santa Clara, Calif).
When not otherwise specified, code implementations were in-house developments based on python 3.6.5 (\url{https://www.python.org}) and on the software modules Numpy, Scipy, and Pytorch 1.1.0.

\subsection*{Model Interpretation}
To reveal the connection between input features (pixels) and the model's final predictions, we back-propagate the loss gradients with respect to their input pixels. 
For all generated gradients, we simply clipped their values to the range of $\pm$ 3 $\times$ standard deviation around their mean value and normalized them to [-1, 1] \cite{tsipras2018robustness}.

To investigate the interaction between learned representations in the deep neural networks and their batch norm layers, we quantified representation similarity via the linear centered kernel alignment (linear CKA).
For a given input and a model, a linear CKA is defined as:
\begin{equation}
	\text{CKA(X, Y)} = \frac{\left \| Y^TX \right \|_F^2}{(\left \| X^TX \right \|_F \left \| Y^TY \right \|_F)}
\end{equation}
where $X$ and $Y$ correspond to a centered Gram matrix of layer activations. 
In particular, in Fig. \ref{fig:representations}, we computed the linear CKA matrix across all 202 radiographs from the internal CheXpert test set. 

\subsection*{Feature Evaluation by Radiologists}
To evaluate the clinical utility of the generated saliency maps for the three models (standard model, adversarially trained model with a single batch norm, and adversarially trained model with the dual batch norm), we randomly chose 100 images from each of the three datasets used in this study (in total 300 images) and let six radiologists assess how useful the map was in guiding a radiologist to the correct diagnosis.
We used a scale from zero, signifying no correlation between the pathology and the saliency map, to five, signifying a map that points clearly and unambiguously to the correct pathology - or pathologies if multiple pathologies were present in the image, see table \ref{Table: rater_score}.
All readers performed the task independently of each other.

\begin{table}[h!]
	\centering
	\captionof{table}{\textbf{Rating standard used for evaluating the diagnostic value of generated saliency maps}}
	\begin{tabular}{cl}
	Score & Diagnostic Rating \\ \hline
	\rowcolor[HTML]{EFEFEF}
	0 & \begin{tabular}[c]{@{}l@{}}no correlation between pathology/ies and hot spot(s) on saliency map.\end{tabular} \\
	1 & highly doubtful low-degree correlation between pathology/ies and hot spot(s) on saliency map. \\
	\rowcolor[HTML]{EFEFEF}
	2 & \begin{tabular}[c]{@{}l@{}}doubtful moderate-degree correlation between pathology/ies and hot spot(s) on saliency map.\end{tabular} \\
	3 & definite partial correlation between pathology/ies and hot spot(s) on saliency map. \\
	\rowcolor[HTML]{EFEFEF}
	4 & \begin{tabular}[c]{@{}l@{}}definite substantial correlation between pathology/ies and hot spot(s) on saliency map.\end{tabular} \\
	5 & \begin{tabular}[c]{@{}l@{}}clear and unambiguous correlation between pathology/ies and hot spot(s) on saliency map.\end{tabular} \\ \hline
	\end{tabular}
	\label{Table: rater_score}
	\end{table}

\subsection*{Statistical Analysis}
For each of the experiments, we calculated the following parameters on the test set: area under the curve (AUC) for the receiver operator characteristic (ROC), sensitivity, and specificity.
The cutoff value for deciding between the presence or non-presence of a pathology was determined by minimizing  $(1- \text{sensitivity})^2 + (1- \text{specificity})^2$ \cite{kniep2019radiomics}.
To assess errors due to sampling of the specific test set and estimate the confidence intervals we employed bootstrap analysis with 10,000 redraws.
The difference in metrics, such as AUC, sensitivity, and specificity, was defined as a $\Delta$metric. 
For the total number of N = 1,000 bootstrapping, models were built after randomly permuting predictions of two classifiers, and metric's differences $\Delta \text{metric}_i$ were computed from their respective scores. 
We obtained the p-value of individual metrics by counting all $\Delta \text{metric}_i$ above the threshold $\Delta$metric.
Statistical significance was defined as P $<$ 0.001.
To determine if differences were significant in the reader studies we employed the Friedman test to test for the presence of differences within the three groups.
If the Friedman test was significant we tested for pairwise differences (i.e. SSM vs. SSBN, SSM vs. SDBN, and SSBN vs. SDBN) employing the Wilcoxon signed-rank test.

\subsubsection*{Author contributions:}
TH and DT devised the concept of the study, DT, SN, MS, MZ, FP, and MH performed the reader tests. TH wrote the code and performed the accuracy studies. TH and DT did the statistical analysis. TH, DT, and VS wrote the first draft of the manuscript. All authors contributed to correcting the manuscript.
\subsubsection*{Disclosures of Conflicts of Interest:}
There are no conflicts of interest to declare.
\subsubsection*{Data and materials availability:} 
This study used four publicly available datasets: 
NIH ChestX-ray14 dataset: \url{https://nihcc.app.box.com/v/ChestXray-NIHCC}; 
Stanford CheXpert dataset: \url{https://stanfordmlgroup.github.io/competitions/chexpert}; kneeMRI dataset: \url{http://www.riteh.uniri.hr/~istajduh/projects/kneeMRI/};
LUNA16 dataset: \url{https://luna16.grand-challenge.org/Data/}
Details of the implementation, as well as the weights of the neural networks after training and the full code producing the results of this paper, are made publicly available under \url{https://github.com/peterhan91/Medical-Robust-Training}. 
All data needed to evaluate the findings in the paper are present in the paper and/or the supplementary material. 
Additional data related to this paper may be requested from the authors.

\bibliography{lib}

\newpage

\section*{SUPPLEMENTARY MATERIALS}
\renewcommand{\thefigure}{S\arabic{figure}}
\renewcommand{\thetable}{S\arabic{table}}
\setcounter{figure}{0}
\setcounter{table}{0}

\begin{table}[h]
	\centering
	\captionof{table}{\textbf{Data characteristics of ChestX-ray8 and CheXpert dataset.}}
	\scalebox{0.9}{
	\begin{tabular}{l|ll}
		& ChestX-ray8 dataset & CheXpert dataset \\ 
		\hline
		\rowcolor{lightgray}
		Number of patient radiographs & 112,120                   & 224,316                   \\
		Number of patients            & 30,805                    & 65,240                    \\
		\rowcolor{lightgray}
		Age, mean (SD),  years        & 46.9 (16.6)               & 60.7 (18.4)               \\
		Percentage of females (\%)    & 43.5\%           		  & 40.6\%           \\
		\rowcolor{lightgray}
		Number of pathology labels    & 8                        & 14                        \\
		Cardiomegaly                  & 2,776                     & 30,092                    \\
		\rowcolor{lightgray}
		Edema                         & 2,303                     & 61,493                    \\
		Effusion                      & 13,317                    & 86,477                    \\
		\rowcolor{lightgray}
		Pneumothorax                  & 5,302                     & 20,401                    \\
		Atelectasis                   & 11,559                    & 59,583                    \\
		\rowcolor{lightgray}
		Consolidation                 & 4,667                     & 12,983                    \\
		Pneumonia                     & 1,431                     & 20,656
	\end{tabular}}
	\caption*{\small
		The seven overlapping labels between the CheXpert and ChestX-ray8 datasets are listed in the table above.
		Note that pathological radiographs seem to be more common in CheXpert dataset due to the differences in the labeling process \cite{irvin2019chexpert}. \\
		Abbreviations: SD, standard deviation.}
	\label{Table: xrays_meta}
\end{table}

\begin{table}[h!]
	\centering
	\captionof{table}{\textbf{Individual and pooled ratings of radiologists in guiding them to the correct pathology on a scale from 0 (useless) to 5 (saliency map points clearly and unambiguously to the correct pathology).}}
	\scalebox{0.9}{
	\begin{tabular}{cccccc}
		& Dataset & Score for SSM & Score for SSBN & Score for SDBN & Friedman test\\ \hline
	\rowcolor[HTML]{EFEFEF}
	& X-ray & $0.32 \pm 0.51$ & $1.56 \pm 0.84$ & $2.12 \pm 1.10$ & $p = 2.1 \cdot 10^{-34}$ \\
	Radiologist 1 (8 years) & MRI & $0.19 \pm 0.39$ & $0.39 \pm 0.57$ & $1.15 \pm 0.92$ & $p = 2.1 \cdot 10^{-21}$ \\
	\rowcolor[HTML]{EFEFEF}
	& CT  & $0.65 \pm 0.50$ & $1.59 \pm 0.73$ & $1.73 \pm 0.83$ & $p = 1.5 \cdot 10^{-33}$ \\ \hline \hline
	\rowcolor[HTML]{EFEFEF}
	& X-ray & $1.53 \pm 1.16$ & $2.86 \pm 0.51$ & $3.30 \pm 0.83$ & $p = 1.1 \cdot 10^{-25}$ \\
	Radiologist 2 (8 years) & MRI & $1.65 \pm 0.61$ & $1.05 \pm 0.78$ & $2.70 \pm 1.22$ & $p = 3.2 \cdot 10^{-21}$ \\
	\rowcolor[HTML]{EFEFEF}
	& CT  & $1.59 \pm 0.83$ & $2.70 \pm 1.08$ & $2.93 \pm 1.18$ & $p = 5.0 \cdot 10^{-23}$ \\ \hline \hline
	\rowcolor[HTML]{EFEFEF}
	& X-ray & $0.29 \pm 0.57$ & $2.47 \pm 1.64$ & $3.10 \pm 2.00$ & $p = 7.9 \cdot 10^{-30}$ \\
	Radiologist 3 (8 years) & MRI & $0.32 \pm 0.62$ & $0.54 \pm 1.21$ & $2.34 \pm 1.97$ & $p = 8.7 \cdot 10^{-20}$ \\
	\rowcolor[HTML]{EFEFEF}
	& CT  & $0.11 \pm 0.31$ & $2.50 \pm 2.13$ & $2.67 \pm 2.14$ & $p = 1.6 \cdot 10^{-22}$ \\ \hline \hline
	\rowcolor[HTML]{EFEFEF}
	& X-ray & $1.03 \pm 1.21$ & $2.49 \pm 1.32$ & $2.52 \pm 1.34$ & $p = 4.8 \cdot 10^{-19}$ \\
	Radiologist 4 (6 years) & MRI  & $0.24 \pm 0.51$ & $1.24 \pm 1.30$ & $2.83 \pm 1.46$ & $p = 2.0 \cdot 10^{-28}$ \\ 
	\rowcolor[HTML]{EFEFEF}
	& CT & $0.37 \pm 0.58$ & $2.92 \pm 1.83$ & $3.10 \pm 1.72$ & $p = 2.7 \cdot 10^{-30}$ \\ \hline \hline
	\rowcolor[HTML]{EFEFEF}
	& X-ray & $0.14 \pm 0.40$ & $1.90 \pm 1.42$ & $2.50 \pm 1.74$ & $p = 4.2 \cdot 10^{-29}$ \\
	Radiologist 5 (5 years) & MRI  & $0.20 \pm 0.49$ & $0.40 \pm 1.09$ & $1.90 \pm 1.71$ & $p = 1.3 \cdot 10^{-20}$ \\ 
	\rowcolor[HTML]{EFEFEF}
	& CT & $0.06 \pm 0.24$ & $2.11 \pm 1.91$ & $2.29 \pm 1.91$ & $p = 2.5 \cdot 10^{-22}$ \\ \hline \hline
	\rowcolor[HTML]{EFEFEF}
	& X-ray & $0.13 \pm 0.34$ & $1.92 \pm 1.45$ & $2.57 \pm 1.76$ & $p = 3.3 \cdot 10^{-29}$ \\
	Radiologist 6 (5 years) & MRI  & $0.09 \pm 0.29$ & $0.81 \pm 1.11$ & $2.11 \pm 1.31$ & $p = 2.6 \cdot 10^{-26}$ \\ 
	\rowcolor[HTML]{EFEFEF}
	& CT & $0.04 \pm 0.20$ & $2.08 \pm 1.85$ & $2.25 \pm 1.92$ & $p = 9.9 \cdot 10^{-23}$ \\ \hline  \hline
	\rowcolor[HTML]{EFEFEF}
	& X-ray & $0.57 \pm 0.94$ & $2.20 \pm 1.33$ & $2.69 \pm 1.56$ & $p = 1.0 \cdot 10^{-160}$ \\
	All radiologists & MRI & $0.49 \pm 0.74$ & $0.74 \pm 1.09$ & $2.17 \pm 1.57$ & $p = 1.7 \cdot 10^{-118}$ \\ 
	\rowcolor[HTML]{EFEFEF}
	& CT  & $0.47 \pm 0.73$ & $2.32 \pm 1.72$ & $2.50 \pm 1.74$ & $p = 7.0 \cdot 10^{-150}$ \\
	\end{tabular}
	}
	\label{table:individualRatings}
\end{table}

\begin{table}[h!]
	\centering
	\captionof{table}{\textbf{Comparison of standard and robust models on Luna16 test set.}}
	\scalebox{0.9}{
	\begin{tabular}{lllllll}
	\hline
	\multicolumn{1}{l|}{\textbf{Prediction}} &
	  \multicolumn{1}{l|}{\textbf{\begin{tabular}[c]{@{}l@{}}ROC-AUC\\ (95\% CI)\end{tabular}}} &
	  \multicolumn{1}{l|}{\textbf{p-value}} &
	  \multicolumn{1}{l|}{\textbf{\begin{tabular}[c]{@{}l@{}}Sensitivity\\ (95\% CI)\end{tabular}}} &
	  \multicolumn{1}{l|}{\textbf{p-value}} &
	  \multicolumn{1}{l|}{\textbf{\begin{tabular}[c]{@{}l@{}}Specificity\\ (95\% CI)\end{tabular}}} &
	  \textbf{p-value} \\ \hline
	  \rowcolor{lightgray}
	\textbf{Tumor malignancy} &  &  &  &  &  &  \\ \hline
	\multicolumn{1}{l|}{Standard model} &
	  \multicolumn{1}{l|}{\begin{tabular}[c]{@{}l@{}}0.952\\ (0.946, 0.959)\end{tabular}} &
	  \multicolumn{1}{l|}{-} &
	  \multicolumn{1}{l|}{\begin{tabular}[c]{@{}l@{}}0.846\\ (0.827, 0.865)\end{tabular}} &
	  \multicolumn{1}{l|}{-} &
	  \multicolumn{1}{l|}{\begin{tabular}[c]{@{}l@{}}0.933\\ (0.923, 0.943)\end{tabular}} &
	  - \\ \hline
	\multicolumn{1}{l|}{\begin{tabular}[c]{@{}l@{}}Robust model,\\ with dual batch norms\end{tabular}} &
	  \multicolumn{1}{l|}{\begin{tabular}[c]{@{}l@{}}0.955\\ (0.949, 0.961)\end{tabular}} &
	  \multicolumn{1}{l|}{0.472} &
	  \multicolumn{1}{l|}{\begin{tabular}[c]{@{}l@{}}0.893\\ (0.878, 0.909)\end{tabular}} &
	  \multicolumn{1}{l|}{0.308} &
	  \multicolumn{1}{l|}{\begin{tabular}[c]{@{}l@{}}0.878\\ (0.864, 0.891)\end{tabular}} &
	  0.275 \\ \hline
	\end{tabular}
	}
	\caption*{\small
	A p-value $<$ 0.001 indicates statistical significance.
	Abbreviations: ROC-AUC, the area under the receiver operating characteristic curve; CI, confidence interval.}
	\label{Table: luna}
	\end{table}

\begin{table}[h!]
	\centering
	\captionof{table}{\textbf{Comparison of standard and robust models on kneeMRI test set.}}
	\scalebox{0.9}{
	\begin{tabular}{lllllll}
	\hline
	\multicolumn{1}{l|}{\textbf{Prediction}} & \multicolumn{1}{l|}{\textbf{\begin{tabular}[c]{@{}l@{}}ROC-AUC\\ (95\% CI)\end{tabular}}} & \multicolumn{1}{l|}{\textbf{p-value}} & \multicolumn{1}{l|}{\textbf{\begin{tabular}[c]{@{}l@{}}Sensitivity\\ (95\% CI)\end{tabular}}} & \multicolumn{1}{l|}{\textbf{p-value}} & \multicolumn{1}{l|}{\textbf{\begin{tabular}[c]{@{}l@{}}Specificity\\ (95\% CI)\end{tabular}}} & \textbf{p-value} \\ \hline
	\rowcolor{lightgray}
	\textbf{Healthy ACL} &  &  &  &  &  &  \\ \hline
	\multicolumn{1}{l|}{Standard model} & \multicolumn{1}{l|}{\begin{tabular}[c]{@{}l@{}}0.824\\ (0.807, 0.841)\end{tabular}} & \multicolumn{1}{l|}{-} & \multicolumn{1}{l|}{\begin{tabular}[c]{@{}l@{}}0.788\\ (0.773, 0.803)\end{tabular}} & \multicolumn{1}{l|}{-} & \multicolumn{1}{l|}{\begin{tabular}[c]{@{}l@{}}0.730\\ (0.701, 0.758)\end{tabular}} & - \\ \hline
	\multicolumn{1}{l|}{\begin{tabular}[c]{@{}l@{}}Robust model,\\ with dual batch norms\end{tabular}} & \multicolumn{1}{l|}{\begin{tabular}[c]{@{}l@{}}0.825\\ (0.808, 0.841)\end{tabular}} & \multicolumn{1}{l|}{0.487} & \multicolumn{1}{l|}{\begin{tabular}[c]{@{}l@{}}0.788\\ (0.773, 0.803)\end{tabular}} & \multicolumn{1}{l|}{0.493} & \multicolumn{1}{l|}{\begin{tabular}[c]{@{}l@{}}0.717\\ (0.688, 0.745)\end{tabular}} & 0.467 \\ \hline
	\rowcolor{lightgray}
	\textbf{Partially injured ACL} &  &  &  &  &  &  \\ \hline
	\multicolumn{1}{l|}{Standard model} & \multicolumn{1}{l|}{\begin{tabular}[c]{@{}l@{}}0.742\\ (0.721, 0.763)\end{tabular}} & \multicolumn{1}{l|}{-} & \multicolumn{1}{l|}{\begin{tabular}[c]{@{}l@{}}0.660\\ (0.625, 0.696)\end{tabular}} & \multicolumn{1}{l|}{-} & \multicolumn{1}{l|}{\begin{tabular}[c]{@{}l@{}}0.735\\ (0.719, 0.750)\end{tabular}} & - \\ \hline
	\multicolumn{1}{l|}{\begin{tabular}[c]{@{}l@{}}Robust model,\\ with dual batch norms\end{tabular}} & \multicolumn{1}{l|}{\begin{tabular}[c]{@{}l@{}}0.741\\ (0.720, 0.761)\end{tabular}} & \multicolumn{1}{l|}{0.482} & \multicolumn{1}{l|}{\begin{tabular}[c]{@{}l@{}}0.626\\ (0.590, 0.661)\end{tabular}} & \multicolumn{1}{l|}{0.393} & \multicolumn{1}{l|}{\begin{tabular}[c]{@{}l@{}}0.731\\ (0.716, 0.747)\end{tabular}} & 0.485 \\ \hline
	\rowcolor{lightgray}
	\textbf{Completely ruptured ACL} &  &  &  &  &  &  \\ \hline
	\multicolumn{1}{l|}{Standard model} & \multicolumn{1}{l|}{\begin{tabular}[c]{@{}l@{}}0.921\\ (0.909, 0.933)\end{tabular}} & \multicolumn{1}{l|}{-} & \multicolumn{1}{l|}{\begin{tabular}[c]{@{}l@{}}0.945\\ (0.915, 0.975)\end{tabular}} & \multicolumn{1}{l|}{-} & \multicolumn{1}{l|}{\begin{tabular}[c]{@{}l@{}}0.802\\ (0.789, 0.815)\end{tabular}} & - \\ \hline
	\multicolumn{1}{l|}{\begin{tabular}[c]{@{}l@{}}Robust model,\\ with dual batch norms\end{tabular}} & \multicolumn{1}{l|}{\begin{tabular}[c]{@{}l@{}}0.918\\ (0.908, 0.929)\end{tabular}} & \multicolumn{1}{l|}{0.499} & \multicolumn{1}{l|}{\begin{tabular}[c]{@{}l@{}}1.000\\ (1.000)\end{tabular}} & \multicolumn{1}{l|}{0.434} & \multicolumn{1}{l|}{\begin{tabular}[c]{@{}l@{}}0.798\\ (0.785, 0.811)\end{tabular}} & 0.494 \\ \hline
	\end{tabular}
	}
	\caption*{\small
	A p-value $<$ 0.001 indicates statistical significance.
	Abbreviations: ACL, anterior cruciate ligament; ROC-AUC, the area under the receiver operating characteristic curve; CI, confidence interval.}
	\label{Table: knee}
	\end{table}

\begin{table}[h!]
	\centering
	\captionof{table}{\textbf{Comparison of standard and robust models on CheXpert test set.}}
	\scalebox{0.9}{
	\begin{tabular}{lllllll}
	\hline
	\multicolumn{1}{l|}{\textbf{Prediction}} & \multicolumn{1}{l|}{\textbf{\begin{tabular}[c]{@{}l@{}}ROC-AUC\\ (95\% CI)\end{tabular}}} & \multicolumn{1}{l|}{\textbf{p-value}} & \multicolumn{1}{l|}{\textbf{\begin{tabular}[c]{@{}l@{}}Sensitivity\\ (95\% CI)\end{tabular}}} & \multicolumn{1}{l|}{\textbf{p-value}} & \multicolumn{1}{l|}{\textbf{\begin{tabular}[c]{@{}l@{}}Specificity\\ (95\% CI)\end{tabular}}} & \textbf{p-value} \\ \hline
	\rowcolor{lightgray}
	\textbf{Cardiomegaly} &  &  &  &  &  &  \\ \hline
	\multicolumn{1}{l|}{Standard model} & \multicolumn{1}{l|}{\begin{tabular}[c]{@{}l@{}}0.798\\ (0.782, 0.814)\end{tabular}} & \multicolumn{1}{l|}{-} & \multicolumn{1}{l|}{\begin{tabular}[c]{@{}l@{}}0.727\\ (0.703, 0.752)\end{tabular}} & \multicolumn{1}{l|}{-} & \multicolumn{1}{l|}{\begin{tabular}[c]{@{}l@{}}0.757\\ (0.741, 0.774)\end{tabular}} & - \\ \hline
	\multicolumn{1}{l|}{\begin{tabular}[c]{@{}l@{}}Robust model,\\ with dual batch norms\end{tabular}} & \multicolumn{1}{l|}{\begin{tabular}[c]{@{}l@{}}0.853\\ (0.841, 0.866)\end{tabular}} & \multicolumn{1}{l|}{0.177} & \multicolumn{1}{l|}{\begin{tabular}[c]{@{}l@{}}0.743\\ (0.718, 0.767)\end{tabular}} & \multicolumn{1}{l|}{0.426} & \multicolumn{1}{l|}{\begin{tabular}[c]{@{}l@{}}0.809\\ (0.794, 0.824)\end{tabular}} & 0.334 \\ \hline
	\rowcolor{lightgray}
	\textbf{Edema} &  &  &  &  &  &  \\ \hline
	\multicolumn{1}{l|}{Standard model} & \multicolumn{1}{l|}{\begin{tabular}[c]{@{}l@{}}0.937\\ (0.929, 0.946)\end{tabular}} & \multicolumn{1}{l|}{-} & \multicolumn{1}{l|}{\begin{tabular}[c]{@{}l@{}}0.881\\ (0.859, 0.904)\end{tabular}} & \multicolumn{1}{l|}{-} & \multicolumn{1}{l|}{\begin{tabular}[c]{@{}l@{}}0.881\\ (0.870, 0.893)\end{tabular}} & - \\ \hline
	\multicolumn{1}{l|}{\begin{tabular}[c]{@{}l@{}}Robust model,\\ with dual batch norms\end{tabular}} & \multicolumn{1}{l|}{\begin{tabular}[c]{@{}l@{}}0.927\\ (0.918, 0.936)\end{tabular}} & \multicolumn{1}{l|}{0.457} & \multicolumn{1}{l|}{\begin{tabular}[c]{@{}l@{}}0.928\\ (0.911, 0.946)\end{tabular}} & \multicolumn{1}{l|}{0.401} & \multicolumn{1}{l|}{\begin{tabular}[c]{@{}l@{}}0.831\\ (0.818, 0.845)\end{tabular}} & 0.361 \\ \hline
	\rowcolor{lightgray}
	\textbf{Consolidation} &  &  &  &  &  &  \\ \hline
	\multicolumn{1}{l|}{Standard model} & \multicolumn{1}{l|}{\begin{tabular}[c]{@{}l@{}}0.918\\ (0.909, 0.927)\end{tabular}} & \multicolumn{1}{l|}{-} & \multicolumn{1}{l|}{\begin{tabular}[c]{@{}l@{}}0.937\\ (0.918, 0.957)\end{tabular}} & \multicolumn{1}{l|}{-} & \multicolumn{1}{l|}{\begin{tabular}[c]{@{}l@{}}0.806\\ (0.792, 0.820)\end{tabular}} & - \\ \hline
	\multicolumn{1}{l|}{\begin{tabular}[c]{@{}l@{}}Robust model,\\ with dual batch norms\end{tabular}} & \multicolumn{1}{l|}{\begin{tabular}[c]{@{}l@{}}0.936\\ (0.928, 0.943)\end{tabular}} & \multicolumn{1}{l|}{0.396} & \multicolumn{1}{l|}{\begin{tabular}[c]{@{}l@{}}0.906\\ (0.883, 0.929)\end{tabular}} & \multicolumn{1}{l|}{0.465} & \multicolumn{1}{l|}{\begin{tabular}[c]{@{}l@{}}0.841\\ (0.828, 0.854)\end{tabular}} & 0.422 \\ \hline
	\rowcolor{lightgray}
	\textbf{Pneumonia} &  &  &  &  &  &  \\ \hline
	\multicolumn{1}{l|}{Standard model} & \multicolumn{1}{l|}{\begin{tabular}[c]{@{}l@{}}0.857\\ (0.837, 0.876)\end{tabular}} & \multicolumn{1}{l|}{-} & \multicolumn{1}{l|}{\begin{tabular}[c]{@{}l@{}}1.000\\ (1.000)\end{tabular}} & \multicolumn{1}{l|}{-} & \multicolumn{1}{l|}{\begin{tabular}[c]{@{}l@{}}0.727\\ (0.713, 0.741)\end{tabular}} & - \\ \hline
	\multicolumn{1}{l|}{\begin{tabular}[c]{@{}l@{}}Robust model,\\ with dual batch norms\end{tabular}} & \multicolumn{1}{l|}{\begin{tabular}[c]{@{}l@{}}0.794\\ (0.768, 0.820)\end{tabular}} & \multicolumn{1}{l|}{0.350} & \multicolumn{1}{l|}{\begin{tabular}[c]{@{}l@{}}1.000\\ (1.000)\end{tabular}} & \multicolumn{1}{l|}{-} & \multicolumn{1}{l|}{\begin{tabular}[c]{@{}l@{}}0.588\\ (0.572, 0.604)\end{tabular}} & 0.304 \\ \hline
	\rowcolor{lightgray}
	\textbf{Atelectasis} &  &  &  &  &  &  \\ \hline
	\multicolumn{1}{l|}{Standard model} & \multicolumn{1}{l|}{\begin{tabular}[c]{@{}l@{}}0.808\\ (0.794, 0.822)\end{tabular}} & \multicolumn{1}{l|}{-} & \multicolumn{1}{l|}{\begin{tabular}[c]{@{}l@{}}0.787\\ (0.765, 0.808)\end{tabular}} & \multicolumn{1}{l|}{-} & \multicolumn{1}{l|}{\begin{tabular}[c]{@{}l@{}}0.756\\ (0.739, 0.773)\end{tabular}} & - \\ \hline
	\multicolumn{1}{l|}{\begin{tabular}[c]{@{}l@{}}Robust model,\\ with dual batch norms\end{tabular}} & \multicolumn{1}{l|}{\begin{tabular}[c]{@{}l@{}}0.818\\ (0.804, 0.832)\end{tabular}} & \multicolumn{1}{l|}{0.426} & \multicolumn{1}{l|}{\begin{tabular}[c]{@{}l@{}}0.800\\ (0.779, 0.821)\end{tabular}} & \multicolumn{1}{l|}{0.455} & \multicolumn{1}{l|}{\begin{tabular}[c]{@{}l@{}}0.772\\ (0.755, 0.789)\end{tabular}} & 0.454 \\ \hline
	\rowcolor{lightgray}
	\textbf{Pneumothorax} &  &  &  &  &  &  \\ \hline
	\multicolumn{1}{l|}{Standard model} & \multicolumn{1}{l|}{\begin{tabular}[c]{@{}l@{}}0.865\\ (0.836, 0.895)\end{tabular}} & \multicolumn{1}{l|}{-} & \multicolumn{1}{l|}{\begin{tabular}[c]{@{}l@{}}0.714\\ (0.636, 0.792)\end{tabular}} & \multicolumn{1}{l|}{-} & \multicolumn{1}{l|}{\begin{tabular}[c]{@{}l@{}}0.795\\ (0.782, 0.808)\end{tabular}} & - \\ \hline
	\multicolumn{1}{l|}{\begin{tabular}[c]{@{}l@{}}Robust model,\\ with dual batch norms\end{tabular}} & \multicolumn{1}{l|}{\begin{tabular}[c]{@{}l@{}}0.791\\ (0.748, 0.834)\end{tabular}} & \multicolumn{1}{l|}{0.330} & \multicolumn{1}{l|}{\begin{tabular}[c]{@{}l@{}}0.569\\ (0.484, 0.655)\end{tabular}} & \multicolumn{1}{l|}{0.178} & \multicolumn{1}{l|}{\begin{tabular}[c]{@{}l@{}}0.995\\ (0.993, 0.997)\end{tabular}} & 0.179 \\ \hline
	\rowcolor{lightgray}
	\textbf{Effusion} &  &  &  &  &  &  \\ \hline
	\multicolumn{1}{l|}{Standard model} & \multicolumn{1}{l|}{\begin{tabular}[c]{@{}l@{}}0.912\\ (0.903, 0.921)\end{tabular}} & \multicolumn{1}{l|}{-} & \multicolumn{1}{l|}{\begin{tabular}[c]{@{}l@{}}0.781\\ (0.758, 0.805)\end{tabular}} & \multicolumn{1}{l|}{-} & \multicolumn{1}{l|}{\begin{tabular}[c]{@{}l@{}}0.870\\ (0.857, 0.882)\end{tabular}} & - \\ \hline
	\multicolumn{1}{l|}{\begin{tabular}[c]{@{}l@{}}Robust model,\\ with dual batch norms\end{tabular}} & \multicolumn{1}{l|}{\begin{tabular}[c]{@{}l@{}}0.931\\ (0.923, 0.939)\end{tabular}} & \multicolumn{1}{l|}{0.383} & \multicolumn{1}{l|}{\begin{tabular}[c]{@{}l@{}}0.797\\ (0.774, 0.820)\end{tabular}} & \multicolumn{1}{l|}{0.429} & \multicolumn{1}{l|}{\begin{tabular}[c]{@{}l@{}}0.891\\ (0.879, 0.903)\end{tabular}} & 0.423 \\ \hline
	\end{tabular}
	}
	\caption*{\small
	A p-value $<$ 0.001 indicates statistical significance.
	Abbreviations: ROC-AUC, the area under the receiver operating characteristic curve; CI, confidence interval.}
	\label{Table: chexpert}
	\end{table}

\subsection*{Reparameterization for adversarial mixture training} \label{Section: repara}
\begin{figure}[h!]
	\centering
	\scalebox{1.0}{
		\includegraphics[trim=0 70 90 0, clip, width=\textwidth]
		{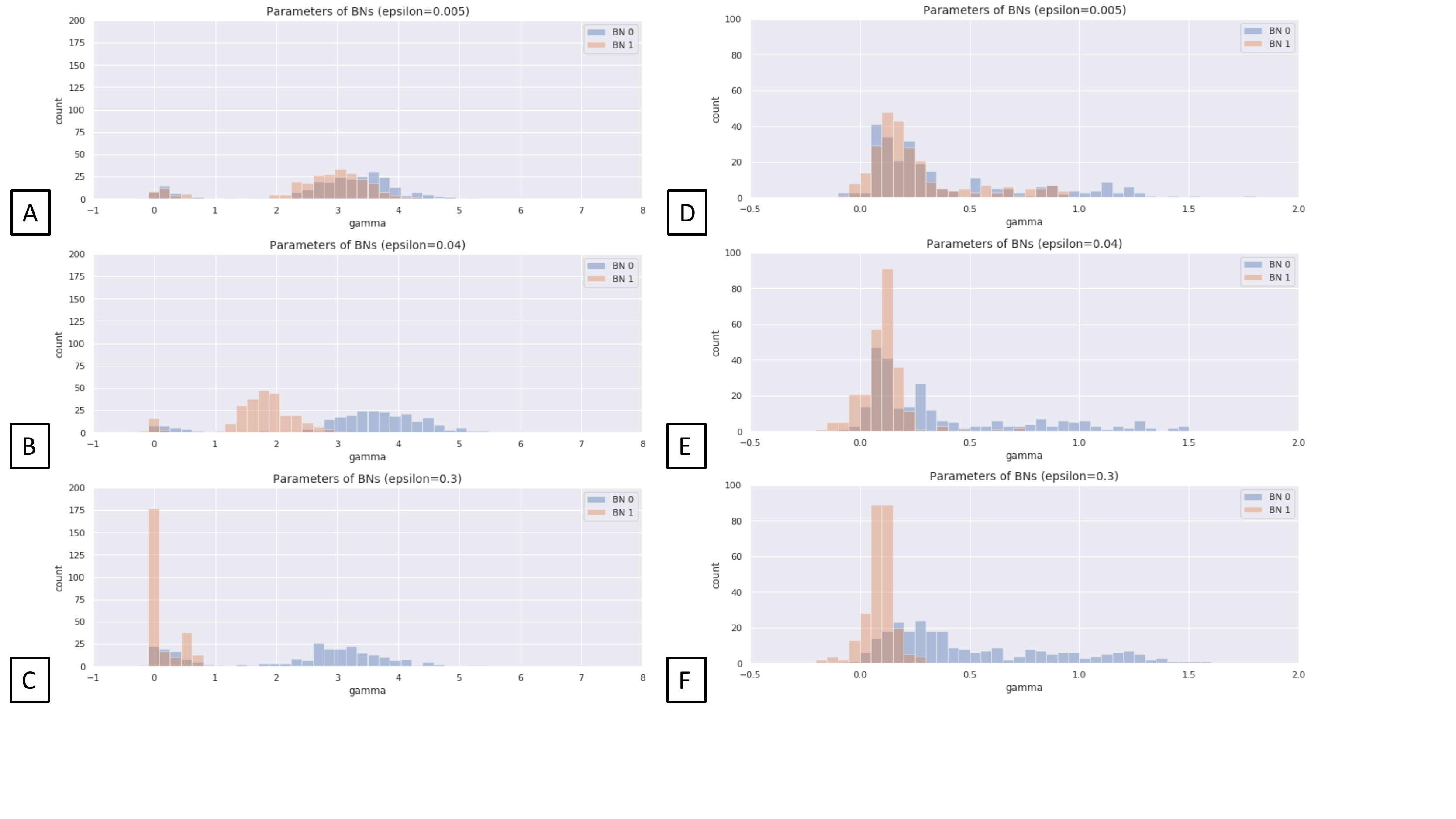}}
	\caption{\label{fig:gamma}\textbf{A smaller magnitude of $\gamma$ was learned when increasing adversarial attack strength during training.}
	Analysis of the reparameterization of batch norms versus adversarial strength $\epsilon =$ 0.005, 0.04, and 0.3.
	A shallow network with just one hidden layer (n=256 neurons and ReLU activation) was trained on both MNIST (A-C) and Luna16 (D-F) datasets. 
	We observe the distribution of adversarial batch $\gamma$ shifts towards zero under stronger adversarial attacks ($\epsilon=0.3$).
	}
\end{figure}

Despite the domain is distinguishable among standard ($x$) and adversarial batches ($x^{\ast}$) \cite{xie2019intriguing}, in our study, we also confirm different reparameterization is applied to those batches. 
By definition, Adversarial batches are difficult to optimize when compared to their standard counterparts.
To investigate, we constructed a simple fully-connected network with one hidden layer followed by two parallel batch norm layers (one for $x$ and one for $x^{\ast}$).
We perform our analysis on Luna16 and MNIST \cite{lecun1998mnist} datasets in combination with different attack strength ($\epsilon$). 
Recall, the Lipschitz constant of the loss, i.e., gradient magnitude $\left \| \hat{g} \right \|$ of a network with batch norms, is governed by reparameterization $\gamma$ \cite{santurkar2018does}. 
Therefore, in adversarial batches, a smaller $\gamma$ is favorable as it corresponds to a smaller Lipschitzness of the loss function and thus facilitates optimization.
As shown in Fig. \ref{fig:gamma} A-C and D-F, the learned $\gamma$  of adversarial batch norms shift to smaller magnitudes when applying stronger attacks, whereas the distribution of standard batch norm remains unchanged. 

\newpage
\subsection*{The influence of perturbation strength on adversarial training}
\begin{figure}[h!]
	\centering
	\scalebox{1.0}{
		\includegraphics[trim=0 80 60 0, clip, width=\textwidth]
		{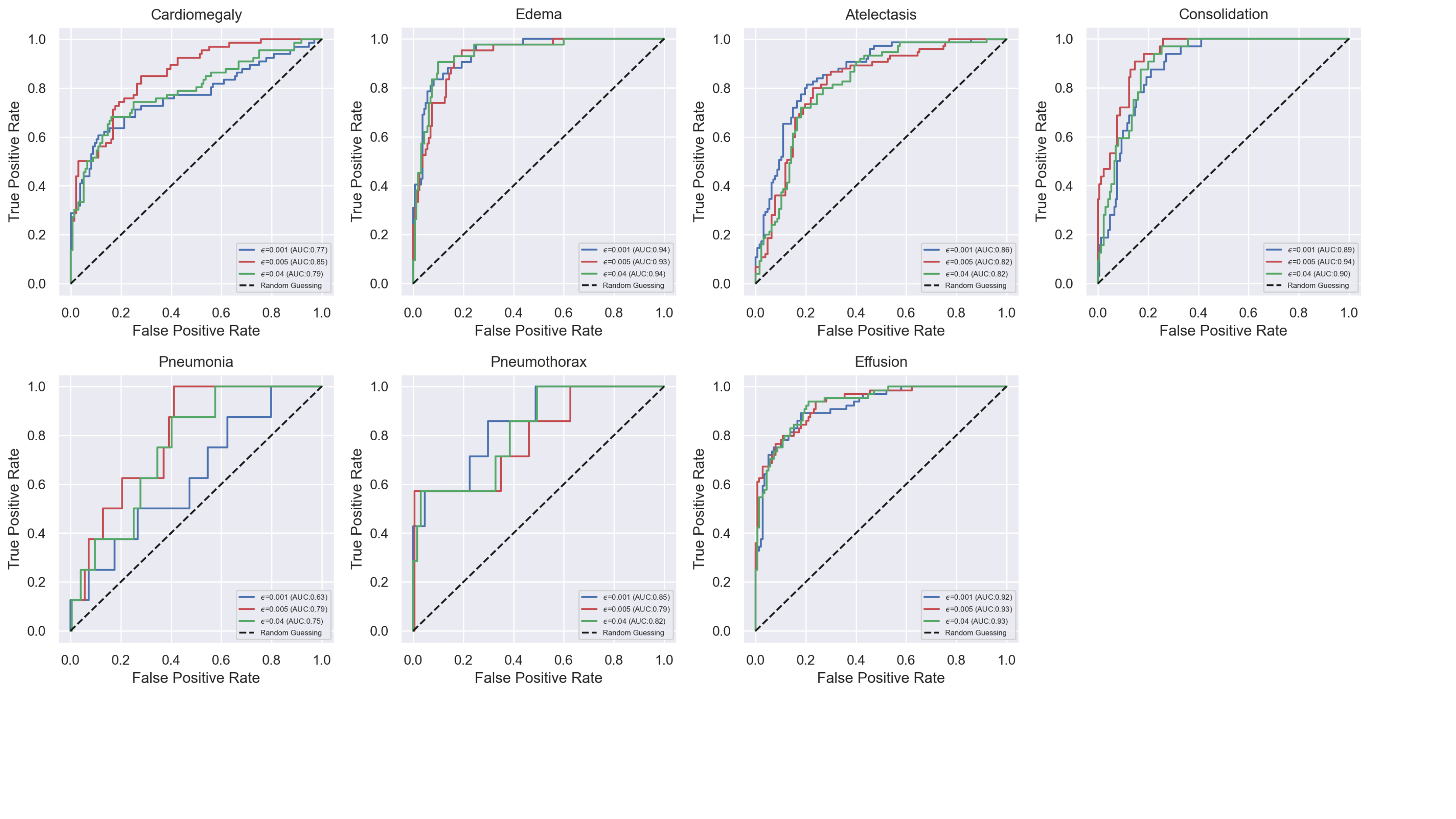}}
	\caption{\label{fig:eps}\textbf{Model performance versus training adversarial strength.}
	We observe that when training with different attack strength, i.e., $\epsilon=$ 0.001, 0.005, and 0.04, the performance of adversarial training with separate batch norms shows little difference. 
	}
\end{figure}

We also investigate how the performance of robust models changes when increasing adversarial attack strength.
Under the setting of training via separate batch norm layers, the model performance is stable as no significant performance decrease was observed when increasing $\epsilon$ from 0.005 to 0.04 (Fig. \ref{fig:eps}). 

\begin{figure}[h!]
	\centering
	\scalebox{1.0}{
		\includegraphics[trim=0 130 0 30, clip, width=\textwidth]
		{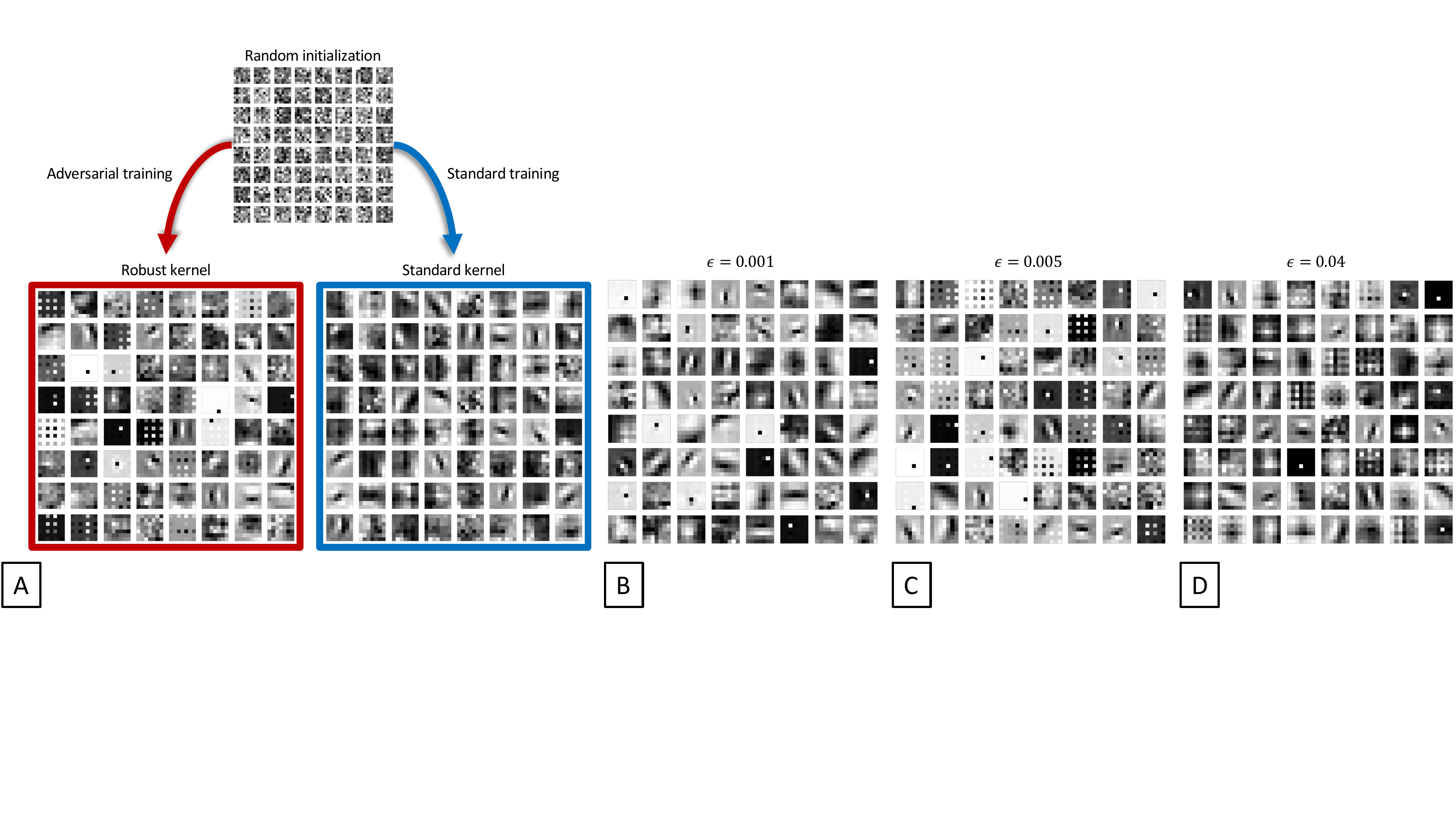}}
	\caption{\label{fig:kernel}\textbf{The influence of training adversarial strength on learned convolutional kernels.}
	Visualization of Conv1 filters (7$\times$7) shows Gabor edge detectors are typically learned through standard training (A). 
	In addition, by adversarial training, robust models learned Dirac-delta functions (kernels) to gain higher robustness \cite{perez2019robust} (B).
	From B to D, we find that checkerboard patterns appear within trained kernels when increasing adversarial strength. 
	}
\end{figure}

However, as visualized in Fig. \ref{fig:kernel}, we observe the general learned Gabor filters (Fig. \ref{fig:kernel} B) are replaced by checkerboard-like filters (Fig. \ref{fig:kernel} C and D) when applying stronger adversarial attacks during training.

\end{document}